\documentclass{article}
\usepackage{ijcai25}

\usepackage{times}
\usepackage{soul}
\usepackage{url}
\usepackage[hidelinks]{hyperref}
\usepackage[utf8]{inputenc}
\usepackage[small]{caption}
\usepackage{graphicx}
\usepackage{amsmath}
\usepackage{amsthm}
\usepackage{booktabs}
\usepackage{algorithm}
\usepackage{algorithmic}
\usepackage[switch]{lineno}
\usepackage{bbding}
\title{Shaping a Stabilized Video by Mitigating Unintended Changes for Concept-Augmented Video Editing}

\author{
Mingce Guo$^1$\and
Jingxuan He$^1$\and
Yufei Yin$^2$\and
Zhangye Wang$^1$\and
Shengeng Tang$^3$\and\\
Lechao Cheng$^3$\textsuperscript{\Envelope}
\\
\affiliations
$^1$Zhejiang University \\
$^2$Hangzhou Dianzi University\\
$^3$Hefei University of Technology \\
\emails
guomingce@zju.edu.cn, 
jingxuan.he@zju.edu.cn,
yinyf@hdu.edu.cn,\\
zywang@cad.zju.edu.cn,
tangsg@hfut.edu.cn, 
chenglc@hfut.edu.cn
}
\usepackage{multirow}
\usepackage{url}
\usepackage{graphicx}
\usepackage{float}
\usepackage{booktabs}
\usepackage{color}
\usepackage{placeins}
\usepackage{amssymb}
\usepackage{xcolor}
\usepackage{algorithm}
\usepackage{algorithmic}
\usepackage{amsmath}

\definecolor{lightblue}{RGB}{0,176,249} 
\usepackage{amsmath,amsfonts,bm}

\def\eqref#1{equation~\ref{#1}}
\def\1{\bm{1}}

\def\vf{{\bm{f}}}

\def\vh{{\bm{h}}}

\def\vv{{\bm{v}}}

\def\vx{{\bm{x}}}

\def\vz{{\bm{z}}}

\def\mA{{\bm{A}}}
\def\mB{{\bm{B}}}

\def\mI{{\bm{I}}}

\def\mK{{\bm{K}}}

\def\mM{{\bm{M}}}

\def\mQ{{\bm{Q}}}

\def\mV{{\bm{V}}}
\def\mW{{\bm{W}}}

\DeclareMathAlphabet{\mathsfit}{\encodingdefault}{\sfdefault}{m}{sl}
\SetMathAlphabet{\mathsfit}{bold}{\encodingdefault}{\sfdefault}{bx}{n}

\def\sD{{\mathbb{D}}}
\def\sF{{\mathbb{F}}}

\begin{document}

\maketitle

\begin{abstract}

Text-driven video editing powered by generative diffusion models holds significant promise for applications spanning film production, advertising, and beyond. However, the limited expressiveness of pre-trained word embeddings often restricts nuanced edits, especially when targeting novel concepts with specific attributes. In this work, we present a novel Concept-Augmented Textual Inversion (CATI) framework that flexibly integrates new object information from user-provided concept videos. By fine-tuning only the V (Value) projection in attention via Low-Rank Adaptation (LoRA), our approach preserves the original attention distribution of the diffusion model while efficiently incorporating external concept knowledge. To further stabilize editing results and mitigate the issue of attention dispersion when prompt keywords are modified, we introduce a Dual Prior Supervision (DPS) mechanism. DPS supervises cross-attention between the source and target prompts, preventing undesired changes to non-target areas and improving the fidelity of novel concepts. Extensive evaluations demonstrate that our plug-and-play solution not only maintains spatial and temporal consistency but also outperforms state-of-the-art methods in generating lifelike and stable edited videos. The source code is publicly available at \href{https://guomc9.github.io/STIVE-PAGE/}{https://guomc9.github.io/STIVE-PAGE/}.

\end{abstract}

\section{Introduction}
Text-driven video editing, powered by generative diffusion models~\cite{ho2020denoising}, ~\cite{song2020denoising},~\cite{rombach2021highresolution}, has emerged as a transformative technology with broad applications in film, art, and advertising~\cite{ho2022imagen},~\cite{hong2022cogvideo}, ~\cite{blattmann2023stable}. Recent advancements, such as Tune-A-Video~\cite{wu2023tune}, FateZero~\cite{qi2023fatezero}, and VideoComposer~\cite{wang2024videocomposer}, have significantly enhanced the ability to edit objects, backgrounds, and styles in video while preserving overall scene consistency through optimized attention mechanisms and spatiotemporal continuity. Despite these successes, existing methods are constrained by the limited expressiveness of CLIP~\cite{radford2021learning} word embeddings, which restricts their ability to perform nuanced edits on targets with specific attributes. Moreover, modifications to the target prompt often disrupt attention mechanisms, leading to inconsistencies in non-target areas before and after editing.

Inspired by Textual Inversion~\cite{gal2022textual}, a feasible approach is to leverage external concept word embeddings, which are optimized within CLIP text encoder~\cite{radford2021learning} while keeping the diffusion model’s parameters frozen. This technique allows the model to incorporate user-provided custom images for guided editing. However, the conventional Textual Inversion faces significant limitations when applied to video editing. Specifically, it lacks the ability to capture novel object information from arbitrary concept videos, resulting in word embeddings with insufficient fidelity to accurately describe target objects. Consequently, directly applying Textual Inversion to one-shot video editing often fails to generate satisfactory results for novel concept pairs, highlighting the need for a more robust and adaptive solution.

To this end, we propose \textbf{Concept-Augmented Textual Inversion} to enable one-shot flexible video editing based on external word embedding and target video. Specifically, we employ cutting-edge LoRA (Low-Rank Adaptation) modules to fine-tune attention value weights, focusing exclusively on the V (Value) projection (we elaborate on the rationale for tuning only V in subsequent sections). This approach effectively maintains the advantages of low VRAM overhead during tuning while preserving the plug-and-play capabilities of the model. In the context of V projection LoRA, our primary objective is to perform inversion while integrating novel object information from arbitrary concept videos for one-shot video editing. The textual inversion process relies on the pre-trained denoising network's established text-image attention probability distribution to achieve accurate target representation. By fine-tuning only the V weights—rather than both Q (Query) and K (Key)—we enable the direct integration of new feature representations while minimizing disruptions to the pre-trained attention distribution. This selective fine-tuning strategy ensures stable training during the early stages, as it suppresses unnecessary changes to the model's foundational attention mechanisms. In addition, we introduce a \textbf{Dual Prior Supervision (DPS)} mechanism, designed to stabilize the generated video by supervising the cross-attention between the source and target prompts. This mechanism addresses the issue of attention dispersion, which often arises when modifications are made to the target prompt. By effectively controlling the attention distribution, DPS significantly enhances the consistency of non-target areas before and after video editing. Furthermore, it enriches the fidelity of the concepts in the edited results, ensuring that the final output maintains both spatial and temporal coherence.

\begin{itemize}
    \item We propose a novel \textbf{Concept-Augmented Textual Inversion (CATI)} approach that reliably captures target attributes from user-provided concept videos, improving the fidelity and flexibility of video editing. 
    \item We introduce a \textbf{Dual Prior Supervision (DPS)} mechanism that stabilizes video generation by supervising cross-attention between source and target prompts. DPS prevents attention dispersion caused by target prompt modifications, significantly improving the consistency of non-target areas before and after editing.
    \item We orchestrate a framework that allows users to extract concepts from custom videos and generate diverse edited videos through concept templates. This approach supports plug-and-play integration with stable diffusion models, enabling efficient and stable video editing.
\end{itemize}

\begin{figure*}[t]
    \centering
    % \vspace{-3mm}
    \includegraphics[width=1.0\linewidth]{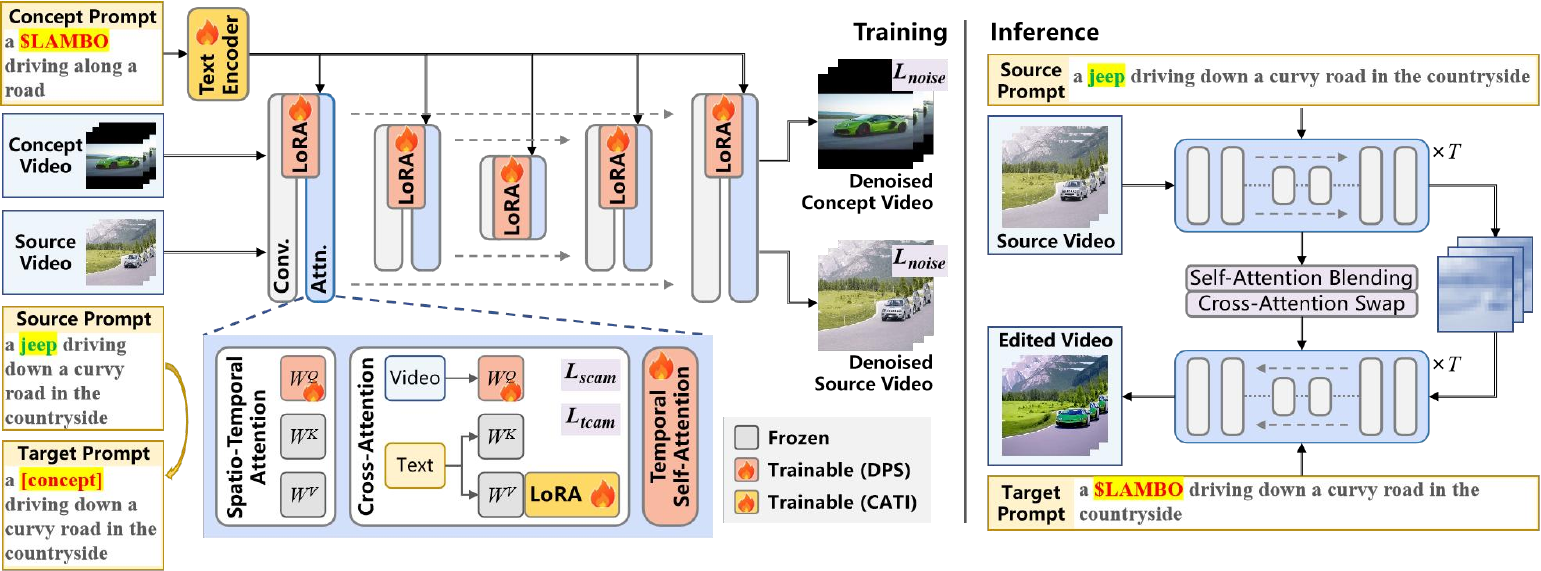}
     % \vspace{-5mm}
    \caption{\textbf{Overview of our training and inference pipelines.} During the training stage, we first adapt the diffusion model to new visual concepts using our introduced Concept-Augmented Textual Inversion (CATI), and then we tune the temporally extended diffusion model with our proposed Dual Prior Supervision (DPS) mechanism to prevent unintended changes in edited videos. During the inference stage, we blend self-attention matrices (Self-Attention Blending) and swap cross-attention matrices (Cross-Attention Swap) to achieve stable video editing. }
    % \vspace{-3mm}
    \label{fig:overview}
\end{figure*}
\section{Related Work}
% \textbf{Text-Driven Video Synthesis.} 
% A series of works based on diffusion models~\cite{ho2020denoising, song2020denoising, rombach2021highresolution} has made significant progress in text-driven image generation. 
% Subsequent efforts~\cite{esser2023structure, wang2023modelscope, blattmann2023stable} further achieve text-driven video generation by extending existing image generation diffusion models. 
% These works commonly inherit the spatial parameters of UNet and fine-tune the newly added temporal modules with large-scale video-text pair datasets to improve inter-frame stability for video synthesis. 
% These works laid a good foundation for video editing with textual descriptions. 
\noindent\textbf{Text-Driven Video Editing.} 
Current approaches for text-driven video editing mainly fall into three categories: fine-tuning video generation models~\cite{zhao2023motiondirector},~\cite{wang2024videocomposer}, fine-tuning image generation models extended with temporal modules~\cite{wu2023tune},~\cite{qi2023fatezero}, and combining NLA~\cite{kasten2021layered} with pre-trained image generation models~\cite{bar2022text2live},~\cite{lee2023shape},~\cite{chai2023stablevideo}. 
% MotionDirector~\cite{zhao2023motiondirector} improves the ability of editing camera and object motions by adding LoRA~\cite{hu2022lora} to the attention modules of the pre-trained Zeroscope~\cite{zeroscope}, strengthening the connection between texts and motions in edited videos. 
% VideoComposer~\cite{wang2024videocomposer} enhances inter-frame consistency by introducing a condition fusion module with spatial and temporal conditions such as motion vectors, depth maps, and sketches. 
% % enhances the stability of objects and actions by incorporating conditional fusion networks for motion vectors, contours, and depth. 
Recent advances have demonstrated various innovative approaches in these categories. For example,~\cite{ku2024anyv2v} employs a pretrained model for diverse video editing tasks, while GenVideo~\cite{singer2025video} utilizes a target-image-aware approach with InvEdit masks to overcome text-prompt limitations. 
% Besides, \cite{singer2025video} introduces the EVE model by distilling pretrained diffusion models.
\cite{bar2022text2live},~\cite{lee2023shape},~\cite{chai2023stablevideo} extract layered neural atlases from video to edit atlases which are further processed to synthesize videos; however, generating a neural atlas demands considerable computational time. 
% NLA~\cite{kasten2021layered} method, 
Recently, Tune-A-Video~\cite{wu2023tune} achieves one-shot video editing with improved inter-frame coherency by updating self-attention with sparse causal attention. 
% builds upon stable-diffusion-v1-4, replacing self-attention with inter-frame sparse causal attention and adding temporal attention mechanisms to enhance the coherence of generated video frames. 
FateZero~\cite{qi2023fatezero} further proposes self-attention blending and incorporates attention control~\cite{hertz2023prompttoprompt} to enhance the ability of editing objects, background, and styles while maintaining scene consistency. For temporal consistency specifically, VidToMe merges self-attention tokens across frames, while \cite{geyer2023tokenflow} leverages inter-frame correspondences to propagate features. In spatial editing, approaches like \cite{ceylan2023pix2video}, \cite{cohen2024slicedit}, \cite{liu2024video} improve results using spatial or temporal attention features in diffusion models.
% FateZero~\cite{qi2023fatezero} further adds temporal convolution layers, proposes self-attention blending, and incorporates attention control methods~\cite{hertz2023prompttoprompt} to enhance the ability to edit scene objects, backgrounds, and styles while maintaining scene consistency. 
For editing targets with specific attributes, it becomes necessary to introduce external word embeddings. 
Our method supports the incorporation of external concept word embeddings. Furthermore, inspired by Tune-A-Video~\cite{wu2023tune} and FateZero~\cite{qi2023fatezero}, we introduce a dual prior supervision mechanism between video latents and word embeddings to enhance scene consistency before and after video editing based on attention control methods. Compared to existing approaches, our method focuses on attention supervision and control mechanisms and operates on a one-shot video editing paradigm.
%It also improves temporal consistency through extended temporal module parameters and enables the flexible integration of external concept objects, while CLIP-based~\cite{radford2021learning} methods above are constrained by finite word embeddings.

\noindent\textbf{Textual Inversion.} 
\cite{gal2022textual} proposes a textual inversion method that optimizes newly added concept word embeddings in the CLIP~\cite{radford2021learning} text encoder, supervised by the latent variable distribution of specific images in the diffusion model. However, using a pre-trained diffusion model for self-supervised text inversion may lead to under-fitting for some specific images due to the finite latent space. Although it's feasible to optimize full parameters of the denosing network in diffusion model with a smaller learning rate simultaneously, or to train it with frozen concept embeddings in the next stage, this process faces issues of easy over-fitting and high storage costs. Our method, building upon textual inversion~\cite{gal2022textual}, attempts to add LoRA~\cite{hu2022lora} to the denosing network, optimizing them simultaneously with concept words at a smaller learning rate, to enhance the text editing capabilities of concept words.

\noindent\textbf{Cross Attention Control and Supervision.} 
Prompt-to-Prompt~\cite{hertz2023prompttoprompt} proposes three attention control methods for stable text-driven image editing based on diffusion models: word swap, refinement, and reweighting. By applying the cross-attention probability map recorded from the original image latent variables and text to the denoising process of original image latent variables and edited text, it has achieved significant success in stable text-driven image editing~\cite{Avrahami_2022_CVPR}, \cite{avrahami2023blendedlatent}. Additionally,~\cite{qi2023fatezero} proposed self-attention blend effectively transfers the stability of text-driven image editing to video editing. Our method, built upon this foundation, introduces external concept words to support editing with higher degrees of freedom.
% However, when performing text-driven editing, whether using existing word embeddings or external concept word embeddings as editing words, there exists a problem of attention dispersion. This means that editing words have non-negligible effects on latent variables other than the editing target. 
Inspired by the work of~\cite{yang2022supervised}, we introduce an attention supervision mechanism to address the issue of dispersed attention in editing words.

\section{Method}

\subsection{Preliminaries}

\textbf{Textual Inversion.}
% Textual inversion~\cite{gal2022textual} learns new embeddings that represent user-provided visual concepts within the textual embedding space. 
% These learned embeddings are then associated with pseudo-words that can be incorporated into novel sentences to achieve text-to-vision editing. 
% The learning process of textual inversion relies on a latent variable diffusion model, which typically consists of an autoencoder and a noise prediction network. 
% For an image $x$, the autoencoder is pretrained such that the encoder $\mathcal{E}$ maps the image to a latent variable $z = \mathcal{E}(x)$, and the decoder $\mathcal{D}$ reconstructs the original image from the latent variable $x \approx \mathcal{D}(z)$. 
% Particularly, textual inversion leverages a CLIP~\cite{radford2021learning} text encoder $c_\theta$ with additional concept words to encode conditional text input $y$. 
% The optimization objective is defined as:
Textual inversion~\cite{gal2022textual} learns new embeddings for user-provided visual concepts in the textual embedding space, associating them with pseudo-words for use in new sentences for text-to-vision editing. The process uses a latent diffusion model, typically with a pretrained autoencoder and a noise prediction network: the encoder $\mathcal{E}$ maps image $x$ to latent $z = \mathcal{E}(x)$, and the decoder $\mathcal{D}$ reconstructs $x \approx \mathcal{D}(z)$. Textual inversion employs a CLIP~\cite{radford2021learning} text encoder $c_\theta$ with added concept words to encode conditional text $y$. The optimization objective is:
\begin{equation}
\label{eq:noise_pred_loss}
\mathcal{L}_{noise}=\mathbb{E}_{z\sim\mathcal{E}(x),y,\epsilon\sim\mathcal{N}(0, 1),t}[\|\epsilon-\epsilon_\theta(z_t, t, c_\theta(y))\|_2^2],
\end{equation}
where $z_t$ is the noised latent at time step $t$, $\epsilon$ is the noise, and $\epsilon_\theta$ is the noise prediction network. 

% For a specific image $x$, a variational autoencoder $\mathcal{E}$ is used to map it to a spatial latent variable $z = \mathcal{E}(x)$. The latent variable diffusion model includes a noise prediction network $\epsilon_\theta$, and a CLIP~\cite{radford2021learning} text encoder $c_\theta$ with additional concept words used for encoding text conditional input $y$. Optimization goal at $t$ time step in diffusion process can then be defined as:
% \begin{equation}
% \mathcal{L}=\mathbb{E}_{z\sim\mathcal{E}(x),y,\epsilon\sim\mathcal{N}(0, 1),t}[\|\epsilon-\epsilon_\theta(z_t, t, c_\theta(y))\|_2^2].
% \end{equation}

\noindent\textbf{Low-Rank Adaption.}
\cite{hu2022lora} proposes an efficient fine-tuning scheme based on matrix low-rank decomposition. For the pre-trained weight $\mW_0 \in \mathbb{R}^{d\times k}$ in the original model, it updates the weight as $\mW = \mW_0 + \Delta \mW$, where $\Delta \mW = \mB\mA$, $\mB \in \mathbb{R}^{d\times r}$, $\mA \in \mathbb{R}^{r\times k}$, and $r \ll min(d, k)$. During the fine-tuning process, the pre-trained weight $\mW_0$ is frozen, while $\mA$ and $\mB$ are trainable parameters. For the forward computation of the original weight $\vh = \mW_0 \vx$, the updated forward computation becomes:
\begin{equation}
\label{eq:lora}
LoRA(\vh) = \mW_0 \vx + \Delta \mW \vx.
\end{equation}

\noindent\textbf{Video Diffusion Models with Temporal Extensions.}
Tune-A-Video~\cite{wu2023tune} introduces Spatio-Temporal Attention (ST-Attn) to replace the original Self-Attention~\cite{vaswani2017attention} in the 2D UNet. 
When calculating the keys $\mK$ and values $\mV$, ST-Attn concatenates latent variables of the first and former frames of the video, leading to the attention result where the current frame attends to both the first and former frames. 
% thereby outputting the attention result of the current frame in relation to both the first and former frames. 
% The specific operations for replacing $\mK, \mV$ in Self-Attention mechanism $Attention(\mQ, \mK, \mV)$ are as follows:
The specific operations for replacing $\mK, \mV$ in Self-Attention are as follows:
\begin{equation}
\label{eq:kv}
    \mK =  \mW^K [\vz_{v_1}; \vz_{v_{i - 1}}], \mV =  \mW^V [\vz_{v_1}; \vz_{v_{i - 1}}],
\end{equation}
where $W^K$ and $W^V$ are projection matrices for key and value respectively, $z_{v_i}$ denotes the latent variable of the $i$-th frame of the video to the current attention layer, and $[\cdot]$ denotes concatenation. 

% 
% NOTE: I rearrange this part to the overview of our framework. 
% 
% While inheriting the pre-trained 2D UNet network parameters from Stable-Diffusion, we add LoRA~\cite{houlsby2019parameter, hu2022lora, yu2023visual} structure temporal convolution layers after the spatial convolution layers, and temporal attention modules with a zero-initialized linear output layer after the cross-attention modules.The outputs of these newly added modules are then residually connected with the outputs of the original modules. 

% These extensions result in a temporally extended 3D UNet~\cite{qi2023fatezero} noise prediction network $\epsilon_{\theta_T}$ for video latents. Optimization goal will be updated to:
% \begin{equation}
% \mathcal{L}=\mathbb{E}_{z\sim\mathcal{E}(x),y,\epsilon\sim\mathcal{N}(0, 1),t}[\|\epsilon-\epsilon_{\theta_T}(z_t, t, c_\theta(y))\|_2^2].
% \end{equation}

% \textbf{Attention Control.}
% In the diffusion process of the source video and source prompt pair, works like Prompt-to-Prompt~\cite{hertz2023prompttoprompt} and FateZero~\cite{qi2023fatezero} save the set  of self and cross attention probability matrices $\sM_{src}$. By employing algorithms such as Word Swap, Phase Adding, and Attention Blending to manipulate the attention probability matrix set $\sM_{tgt}$ during the diffusion process of the source video and target prompt, they achieve stable image and video editing.

\subsection{Stabilized Text-Driven Video Editing}
Our training and inference pipelines are shown in Fig.~\ref{fig:overview}. We use a UNet initialized from Stable Diffusion’s pre-trained 2D UNet~\cite{rombach2021highresolution} as the noise predictor. To handle 3D video inputs, we replace spatial self-attention layers with ST-Attn (Eq.\ref{eq:kv}). Following FateZero~\cite{qi2023fatezero}, we add LoRA-based temporal convolution layers after spatial convolutions, and temporal self-attention with zero-initialized linear output after cross-attention. These new modules are residually connected to the originals.

% While inheriting the pre-trained 2D UNet network parameters from Stable-Diffusion, we add LoRA~\cite{houlsby2019parameter, hu2022lora, yu2023visual} structure temporal convolution layers after the spatial convolution layers, and temporal attention modules with a zero-initialized linear output layer after the cross-attention modules.The outputs of these newly added modules are then residually connected with the outputs of the original modules. 

Our approach for stabilized text-driven video editing has two learning phases. 
In the first phase, we introduce Concept-Augmented Textual Inversion (CATI) to adapt the diffusion model to new visual concepts. 
In the second phase, we tune partial parameters of the temporally extended diffusion model to suppress unintended changes in edited videos by calibrating cross-attention results. 

\noindent\textbf{Concept-Augmented Textual Inversion.}
Textual inversion~\cite{gal2022textual} learns to represent a specific set of user-provided images with pseudo-words in the latent space, offering an intuitive way for natural language-guided image editing. 
We incorporate this technique into our framework to facilitate video editing. 
However, due to the self-supervised nature within the limited latent space of the pre-trained diffusion model, the vanilla textual inversion often results in varied performance in terms of quality and efficiency for different image sets, requiring meticulous adjustments for learning rates and iteration counts. 

% When performing textual inversion~\cite{gal2022textual} on a specific set of user-input images, and then using pre-trained concept words to participate in text-driven diffusion models for image editing, 
% there are limitations due to the self-supervised nature within the limited latent space of the pre-trained diffusion model. This results in differences in quality and efficiency of textual inversion for different image sets, requiring adjustments to learning rates and iteration counts to adapt, or fine-tuning the diffusion model with a smaller learning rate to reduce iteration time costs. 

To alleviate this issue, we draw inspiration from existing parameter-efficient fine-tuning techniques and propose adding LoRA modules~\cite{hu2022lora} to the value projection parameters in the cross-attention layers of the UNet. 
Consequently, the values $\mV$ are updated to $LoRA(\mV)$ according to Eq.~(\ref{eq:lora}). 
The rationale behind our approach is that we aim to enhance the expressiveness of the pre-trained diffusion model by slightly adjusting its capacity to accommodate new visual concepts while preserving its original generation capability. 
Besides, inserting LoRA modules not only augments textual inversion with low storage overhead but also maintains a plug-and-play characteristic during inference. 
% \textcolor{red}{Textual inversion process relies on the pre-trained denoising network established text-image attention probability distribution to achieve accurate target representation. In this context, fine-tuning only the $\mV$ weights instead of $\mQ$ and $\mK$ allows new feature representations to be directly integrated while suppressing changes towards the pre-trained attention distribution to enable stable training during the early stages.}
Textual inversion process relies on the pre-trained denoising network established text-image attention probability distribution to achieve accurate target representation. In this context, fine-tuning only the $\mV$ weights instead of $\mQ$ and $\mK$ allows new feature representations to be directly integrated while suppressing changes towards the pre-trained attention distribution to enable stable training during the early stages.

We train the concept-word embeddings of textual inversion and the weight parameters of LoRA modules in an end-to-end manner (see orange blocks in Fig.~\ref{fig:overview}), where the learning rate for LoRA parameters is relatively smaller than that for concept-word embeddings to avoid over-fitting. 
Denote the noise prediction network with LoRA modules loaded on value projection parameters as $\epsilon_{\theta_L}$, the optimization objective of concept-augmented textual inversion is then updated from Eq.~(\ref{eq:noise_pred_loss}) to the following: 
\begin{equation}
\mathcal{L}_{noise}=\mathbb{E}_{z\sim\mathcal{E}(x),y,\epsilon\sim\mathcal{N}(0, 1),t}[\|\epsilon-\epsilon_{\theta_L}(z_t, t, c_\theta(y))\|_2^2].
\end{equation}

% To alleviate this issue, we combine existing fine-tuning techniques and attempt to add LoRA~\cite{hu2022lora} modules to the Value weight parameters in the cross-attention layers of the UNet in the diffusion model, specifically replacing the output $\mV$ of the Value module with $LoRA(\mV)$. We control the concept word embeddings and the associated learnable LoRA weight parameters to iterate at different learning rates. This approach improves the efficiency of textual inversion with low additional storage overhead while maintaining plug-and-play characteristics during inference. After inserting the LoRA module, the noise prediction function $\epsilon_\theta$ is updated to $\epsilon_{\theta_L}$, and the optimization objective is then replaced to:
% \begin{equation}
% \mathcal{L}=\mathbb{E}_{z\sim\mathcal{E}(x),y,\epsilon\sim\mathcal{N}(0, 1),t}[\|\epsilon-\epsilon_{\theta_L}(z_t, t, c_\theta(y))\|_2^2].
% \end{equation}

\begin{figure}[t]
    \centering
    % \vspace{-3mm}
    \includegraphics[width=1.0\linewidth]{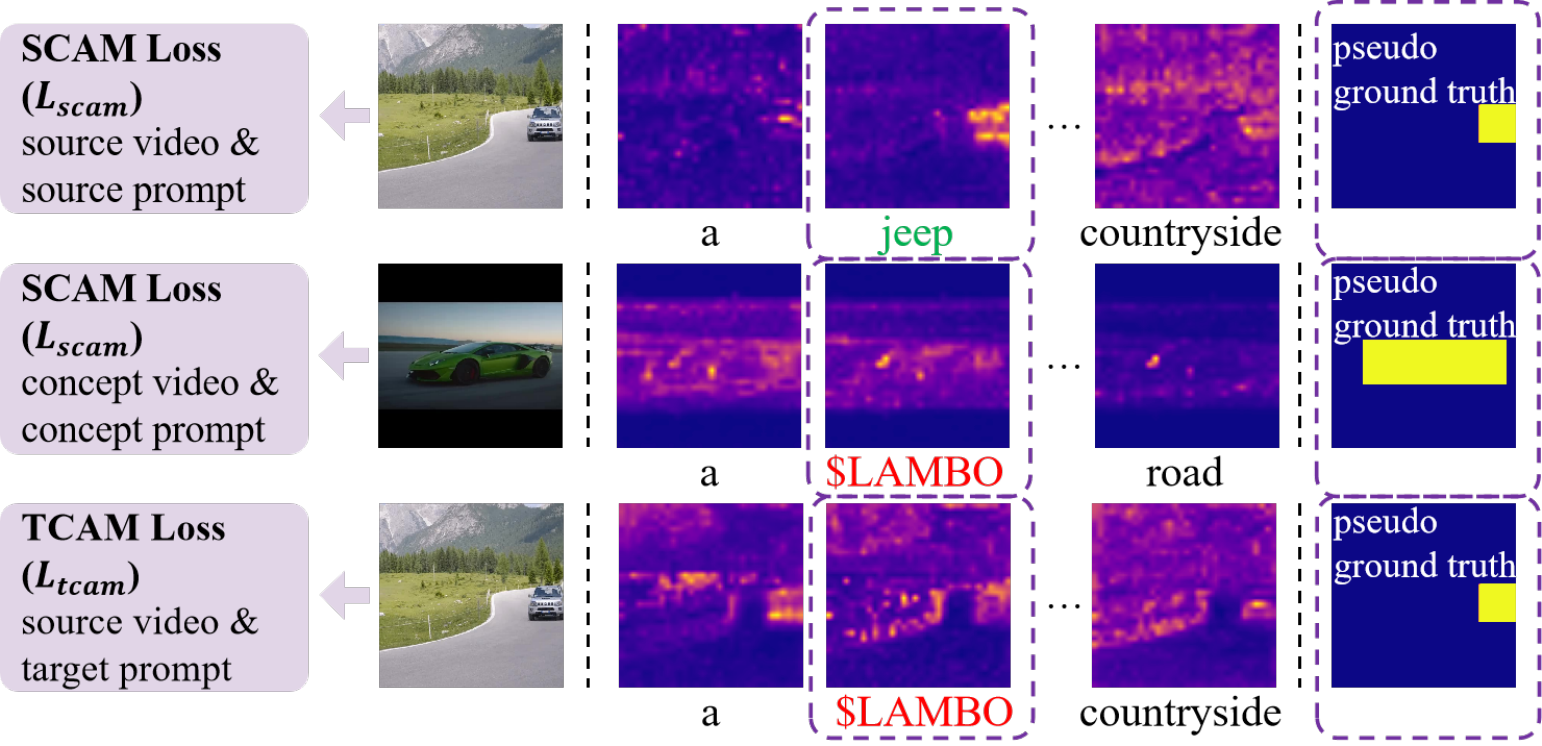}
    % \vspace{-5mm}
    \caption{\textbf{Visualization of the dual prior supervision mechanism.} Each row displays a video frame, a set of cross-attention maps between this video frame and prompt words, and a pseudo ground truth mask. The \textit{scam} loss and \textit{tcam} loss are computed between relevant words and pseudo masks to reduce unintended changes. }
    % \vspace{-5mm}
    \label{fig:losses}
\end{figure}

% \textbf{Tune Video Diffusion Model with Dual Prior Supervision.}
\noindent\textbf{Model Tuning with Dual Prior Supervision.}\label{sec:dps}
After learning concept-augmented textual inversion, we adapt and tune the video diffusion model for text-driven video editing in line with the paradigm of few-shot learning. 
Specifically, we learn the LoRA-structured temporal convolution layers, the query projection weights within spatio-temporal attention layers and cross-attention layers, and the temporal self-attention layers (see red blocks in Fig.~\ref{fig:overview}). 
These parameters are selected for updates during training due to their strong relevance to the temporal modeling of 3D videos. 
To attain more stable and higher quality editing results, we tried directly integrating existing attention control techniques~\cite{hertz2023prompttoprompt} in an early attempt; however, we found that when applying text-driven video editing types such as word swap, the dispersion phenomenon of cross-attention between text embeddings and video latents leads to reduced stability in editing results, which is shown in Fig~\ref{fig:abla-full-supvis}. 
To address this challenge, we propose a dual prior supervision mechanism, which includes a source cross-attention mask (\textit{scam}) loss and a target cross-attention mask (\textit{tcam}) loss. 

% % Question: why cite a zero-shot work when mentioning fine-tuning the model???
% For text-driven editing of few-shot video data, fine-tuning the text-driven video generation model~\cite{qi2023fatezero}, which is a temporal extension of Stable-diffusion, can be used to achieve this. 
% Combined with attention control techniques like~\cite{hertz2023prompttoprompt}, more stable and higher quality editing effects can be obtained. 
% However, we found that when using text-driven video editing types such as word swap, the cross-attention dispersion of editing words on video latent variables leads to reduced stability in editing results. Therefore, we propose a method of supervising cross-attention using masks generated from pseudo-labels.

The \textit{scam} loss is designed to reduce the attention influence of the words to be replaced in the source prompt on irrelevant frame areas (see the first row in Fig.~\ref{fig:losses}). 
It is also applied to modulate attention between concept words and concept videos (see the second row in Fig.~\ref{fig:losses}). 
Specifically, for $K$ cross-attention layers in the UNet, we record cross-attention matrices $\mathbf{M}_{s}$ between the words and the video frame latents in each cross-attention layer. 
To obtain ground truth for optimization, we use an off-the-shelf object detection network OWL-ViT~\cite{minderer2022simple} to localize objects in video frames and generate corresponding binary pseudo-labels ${\mathbf{M}}^{\text{gt}}_{s}$. 
We further apply max pooling to generate $K$ pseudo-labels, each with a designated resolution $P_k$. 
The loss is then calculated as the mean absolute loss on irrelevant areas: 
\begin{small}
    \begin{equation}
        \mathcal{L}_{scam} = \frac{1}{K} \sum_{k=1}^{K} \sum_{i=1}^{P_k} \left [ \| \mathbf{M}^{\text{gt}}_{s,k,i} - \mathbf{M}_{s,k,i} \| \cdot ( 1 - \mathbf{M}^{\text{gt}}_{s,k,i} ) \right ].
    \end{equation}
\end{small}
% We then apply max pooling to the generated pseudo-labels to obtain a set of masks $\sM_s^{gt}$ with a set of \textcolor{red}{different resolutions $\sP$}, of length $K$. To supervise the attention distribution of the prompt words to be replaced in the source video and reduce their attention influence on irrelevant areas, we propose the calculation method for the scam (source cross attention mask) loss as follows, where $\overline{\vm}_{s,k}\in \sM_s^{gt}$, $\vm_{s,k}\in\sM_s$ and $p_k\in\sP$.
% \begin{equation}
%     \mathcal{L}_{scam}=\frac{1}{K} \textcolor{red}{\sum_{k=0}^{K}\sum_{i=0}^{p_k}} [\| \overline{m}_{s,k,i}-m_{s,k,i}\| \cdot (1-\overline{m}_{s,k,i})].
% \end{equation}

The \textit{tcam} loss is introduced to diminish the attention influence of the target words in the edited prompt to further promote the consistency of irrelevant areas before and after video editing (see the third row in Fig.~\ref{fig:losses}). 
Similar to the \textit{scam} loss, we obtain cross-attention matrices $\mathbf{M}_{t}$ and pseudo-labels $\mathbf{M}^{\text{gt}}_{t}$ between the target words in the edited prompt and the video frame latents. 
The loss is computed as:
\begin{small}
    \begin{equation}
        \mathcal{L}_{tcam} = \frac{1}{K} \sum_{k=1}^{K} \sum_{i=1}^{P_k} \left [ \| \mathbf{M}^{\text{gt}}_{t,k,i}-\mathbf{M}_{t,k,i} \| \cdot ( 1 - \mathbf{M}^{\text{gt}}_{t,k,i} ) \right].
    \end{equation}
\end{small}
% To supervise the attention distribution between the target words in the edited prompt and the video to be edited, it is also necessary to calculate the tcam(target cross attention mask) loss in each iteration to reduce their attention influence on irrelevant areas. Similarly, we obtain the cross-attention probability set $\sM_t$ and mask set $\sM_t^{gt}$ between the target words in the edited text and the latent variables of the target video. The calculation method is as follows, where $\overline{\vm}_{t,k}\in \sM_t^{gt}$, $\vm_{t,k}\in\sM_t$ and $p_k\in\sP$.
% \begin{equation}
%     \mathcal{L}_{tcam}=\frac{1}{K}\sum_{k=0}^{K}\sum_{i=0}^{p_k} [\| \overline{m}_{t,k,i}-m_{t,k,i}\| \cdot (1-\overline{m}_{t,k,i})].
% \end{equation}

Let the trainable parameters during the model tuning phase be denoted as $\epsilon_{\theta_{T}}$. 
The noise prediction loss $\mathcal{L}_{noise}$ is then obtained by substituting $\epsilon_{\theta}$ in Eq.~(\ref{eq:noise_pred_loss}) with $\epsilon_{\theta_{T}}$. 
Given $\alpha$ and $\beta$ as the weighting coefficients for our proposed \textit{scam} loss and \textit{tcam} loss, respectively, the total loss for model tuning with dual prior supervision is formulated as: 
\begin{equation}
\mathcal{L}=\mathcal{L}_{noise}+\alpha \mathcal{L}_{scam}+\beta \mathcal{L}_{tcam}.
\label{eq:supvis}
\end{equation}

% Additionally, to complement the concept-augmented textual inversion for user-input custom target videos, we insert pre-trained LoRA parameters into the noise prediction network $\epsilon_{\theta_{T}}$ of the original temporally extended video generation model to obtain $\epsilon_{\theta_{TL}}$. We then \textcolor{red}{use masks} to supervise the input training source videos.
% \begin{equation}
% \mathcal{L}_{noise}=\mathbb{E}_{z\sim\mathcal{E}(x),y,\epsilon\sim\mathcal{N}(0, 1),t}[\|\epsilon-\epsilon_{\theta_{TL}}(z_t, t, c_\theta(y))\|_2^2].
% \end{equation}
% For the trainable parameters in the noise prediction network $\epsilon_{\theta_{TL}}$, the loss function combining the noise prediction loss and the attention supervision loss is as follows, where $\alpha$ and $\beta$ are the weight parameters for the scam loss and tcam loss respectively.
% \begin{equation}
% \mathcal{L}=\mathcal{L}_{noise}+\alpha \mathcal{L}_{scam}+\beta \mathcal{L}_{tcam}.
% \label{eq:supvis}
% \end{equation}

\noindent\textbf{Inference.}
% As shown in Fig.~\ref{fig:overview}, the inference pipeline involves an inversion stage using the source text prompt, and an editing stage using the modified text prompt. 
% We cache the intermediate self-attention matrices and cross-attention matrices at each time step during inversion. 
% These matrices are then leveraged to manipulate attention during editing. 
% Specifically, we blend self-attention matrices to retain the semantic layout following FateZero~\cite{qi2023fatezero} (Self-Attention Blending), and swap cross-attention matrices between the changed words and video latents similar to Prompt-to-Prompt~\cite{hertz2023prompttoprompt} (Cross-Attention Swap). 
As shown in Fig.\ref{fig:overview}, the pipeline consists of an inversion stage with the source prompt and an editing stage with the modified prompt. During inversion, we cache self- and cross-attention matrices at each step, which are later used to control attention in editing. Specifically, we blend self-attention matrices to preserve semantic layout~\cite{qi2023fatezero}, and swap cross-attention matrices for changed words and video latents~\cite{hertz2023prompttoprompt}.
\section{Experiments}

\subsection{Settings and Datasets}
Our experiments are conducted on a machine equipped with an NVIDIA GeForce RTX 4090. During the concept augmented textual inversion stage, we set the learning rate for CLIP~\cite{radford2021learning} word embeddings to $1\times 10^{-3}$, and the learning rate for LoRA modules inserted into the UNet to $1\times 10^{-5}$, with the number of training steps set to $5000$. %Additionally, in this stage, we augment the length of training frames from the concept video to prevent the inversion process from over-fitting to a fixed frames length. Specifically, We set the length of training frames per iteration within the range $[4, 8]$, and randomly sample sequences of frames of this length from the concept videos to obtain training frames. 
Additionally, we randomly sample frame numbers within the range $[4, 8]$ from the concept video during training, to prevent the inversion process from over-fitting to a fixed frame number. 
For the video diffusion model fine-tuning stage, we empirically set $\alpha=0.1$ and $\beta=0.1$ in Eq.~(\ref{eq:supvis}). The training steps above all use AdamW~\cite{loshchilov2017decoupled} optimizer. In the inference stage of video editing, the guidance scale is set to $12.5$, the number of DDIM Inversion steps is $T=50$, and the self-attention blending and cross-attention swap steps are within the interval $[0, 0.7 T]$. To evaluate our proposed method, we used a portion of the DAVIS~\cite{Caelles_arXiv_2019} dataset and clip videos from the internet to construct video editing pairs, either with or without concept videos.

% useless
% We compare the video editing results of Tune-A-Video~\cite{wu2023tune}, FateZero~\cite{qi2023fatezero}, MotionDirector~\cite{zhao2023motiondirector}, and RAVE~\cite{kara2024rave} methods in scenarios both including and excluding the concept. For MotionDirector, we utilized its supported temporal path and spatial path, training them using the source video and the concept video, respectively.

% \subsection{Datasets}

\subsection{Metrics}

\textbf{Frame Consistency.} 
% To compare the coherence of the video frames $\sF$, we refer to the metric used in~\cite{wu2023tune, hessel2021clipscore}, which construct a set $\sD$ from the vector pairs $(\vv_i, \vv_j)$ encoded by the CLIP visual encoder~\cite{radford2021learning} for any two different frames $\vf_i, \vf_j\in \sF, \vf_i \ne \vf_j$ and calculates the average cosine distance $d$ in Eq.~(\ref{eq:fc}). 
To compare the coherence of the video frames $\sF$, we refer to the metric used in~\cite{wu2023tune}, ~\cite{hessel2021clipscore}, which calculates the average cosine distance $d$ between features $(\vv_i, \vv_j)$  of each two different frames $(\vf_i, \vf_j)$ encoded by the CLIP visual encoder~\cite{radford2021learning}, as Eq.~(\ref{eq:fc}). Here, $\vf_i, \vf_j\in \sF$, $\vf_i \ne \vf_j$, and $\sD$ denotes the set of the vector pairs $(\vv_i, \vv_j)$. 
\begin{equation}
    d=\frac{1}{|\sD|}\sum_{(\vv_i,\vv_j)\in \sD}\frac{\vv_i \cdot \vv_j}{\|\vv_i\| \|\vv_j\|}.
    \label{eq:fc}
\end{equation}

\begin{figure*}[!t]
    \centering
    % \vspace{-10mm}
    \includegraphics[width=0.95\textwidth]{figures/qualitative.pdf}
    % \vspace{-3mm}
    \caption{\textbf{Video generation with (\textbf{Setting I}) and without (\textbf{Setting II}) concept pairs.} The top row of the figure contains the concept video with its prompt. The second row is the source video frames coupled with prompts that need to be edited. The rows below show the editing results of the source video using the editing prompt for~\protect\cite{wu2023tune},~\protect\cite{qi2023fatezero},~\protect\cite{zhao2023motiondirector},~\protect\cite{kara2024rave} and our method, respectively, in which words with "\$" ahead mean concept words, and the same for subsequent results.}
    \label{fig:qual}
    % \vspace{-6mm}
\end{figure*}

\begin{table}
    % \centering
    \resizebox{0.95\linewidth}{!}{
    \begin{tabular}{lccc}
        \toprule
        \begin{small}Methods\end{small} & \begin{small}M-PSNR $\uparrow$\end{small} & \begin{small}Concept Cons. $\uparrow$\end{small} & \begin{small}Frame Cons. $\uparrow$\end{small}\\
        \midrule
        \begin{small}Tune-A-Video\end{small} & 14.70 & 0.6982 & 0.9399 \\
        \begin{small}FateZero\end{small} & 17.08 & 0.6822 & 0.9413 \\
        \begin{small}MotionDirector\end{small} & 12.73 & 0.7222 & 0.9452 \\
        \begin{small}RAVE\end{small} & 17.39 & 0.6990 & 0.9379 \\
        \noalign{\vskip 0.4ex}
        \hline
        \noalign{\vskip 0.4ex}
        \begin{small}Ours\end{small} & \textbf{19.71} & \textbf{0.7642} & \textbf{0.9472}\\
        \bottomrule
    \end{tabular}}
    % \vspace{-2mm}
    \caption{Quantitative results of video editing \textbf{\textit{w/}} concept video.}
    % \vspace{-2mm}
    \label{tab:quan-concept}
\end{table}

\begin{table}
    \centering
    \resizebox{0.7\linewidth}{!}{
    \begin{tabular}{lcc}
        \toprule
        Methods  & M-PSNR $\uparrow$ & Frame Cons. $\uparrow$ \\
        \midrule
        Tune-A-Video & 15.72 & 0.9397 \\
        FateZero & 19.42 & 0.9246 \\
        MotionDirector & 16.86 & 0.9403 \\
        RAVE & 16.20 & 0.9306 \\
        \noalign{\vskip 0.4ex}
        \hline
        \noalign{\vskip 0.4ex}
        Ours & \textbf{22.10} & \textbf{0.9405} \\
        \bottomrule
    \end{tabular}}
    % \vspace{-2mm}
    \caption{Quantitative results of video editing \textbf{\textit{w/o}} concept video.}
    % \vspace{-2mm}
    \label{tab:quan-no_concept}
\end{table}

\noindent\textbf{Masked Peek-Signal-Noise Ratio.}
To compare the stability of the video non-target areas before and after target editing, we design a Masked Peak Signal-to-Noise Ratio (\textbf{M-PSNR}) metric. We use the OWL-ViT~\cite{minderer2022simple} open-vocabulary object detection model with text pseudo-labels to estimate the bounding box mask $\mM$ of the edited target. We then compare the average peek-signal-noise ratio of the original video frames and the edited video frames after applying this mask. The calculation formula for the specific function $f$ for the Mean Squared Error (MSE) used as input is as follows, where $\mM\in \mathbb{R}^{H\times W}$, $\mI^{s}\in \mathbb{R}^{H\times W \times C}$, and $\mI^{e}\in \mathbb{R}^{H\times W \times C}$ refer to the mask value, the frame pixel value of video before and after editing, respectively.
\begin{tiny}
    \begin{equation}
        % f(\mI^{s}, \mI^{e}, \mM)=\frac{1}{C}\sum_{k=0}^{C}\frac{\sum\limits_{i=0}^{H}\sum\limits_{j=0}^{W}(I_{i,j,k}^{s}-I_{i,j,k}^{e})^2(1-M_{i,j})}{\sum\limits_{i=0}^{H}\sum\limits_{j=0}^{W}(1-M_{i,j})}.
         % f(\mI^{s}, \mI^{e}, \mM)=\frac{1}{C}\sum_{k=0}^{C}\frac{\sum_{i=0}^{H}\sum_{j=0}^{W}(I_{i,j,k}^{s}-I_{i,j,k}^{e})^2(1-M_{i,j})}{\sum_{i=0}^{H}\sum_{j=0}^{W}(1-M_{i,j})}.
         % f(\mI^{s}, \mI^{e}, \mM)=\frac{1}{C}\sum_{k\in C}\frac{\sum_{i\in H}\sum_{j\in W}(I_{i,j,k}^{s}-I_{i,j,k}^{e})^2(1-M_{i,j})}{\sum_{i\in H}\sum_{j\in W}(1-M_{i,j})}.
         f(\mI^{s}, \mI^{e}, \mM)=\frac{1}{C}\frac{\sum_{k\in C}\sum_{i\in H}\sum_{j\in W}(I_{i,j,k}^{s}-I_{i,j,k}^{e})^2(1-M_{i,j})}{\sum_{i\in H}\sum_{j\in W}(1-M_{i,j})}.
    \end{equation}
\end{tiny}

\noindent\textbf{Concept Consistency.}
We employ a multi-step approach to evaluate the correlation between the video editing results guided by the concept video and the concept video itself while minimizing interference in non-target areas. First, we use a pre-trained OWL-ViT~\cite{minderer2022simple} model in conjunction with pseudo-label prediction to generate object masks for both videos. We then extract pixel segments of the target objects from both the edited video and the concept video. Finally, we leverage the CLIP model to predict visual encoding vectors for these extracted segments and calculate the average cosine similarity between them. 

% \begin{figure*}[t]
%     \centering
%     \vspace{-3mm}
%     \includegraphics[width=0.9\textwidth]{figures/qualitative.pdf}
%     \vspace{-2mm}
%     \caption{\textbf{Video generation with (\textbf{Setting I}) and without (\textbf{Setting II}) concept pairs.} The top row of the figure contains the concept video with its prompt in the left part, and comparison settings in the right part. The second row is the source video frames to be edited and its prompt. The rows below show the editing results of the source video using the editing prompt, for \cite{wu2023tune}, \cite{qi2023fatezero}, \cite{zhao2023motiondirector}, \cite{kara2024rave} and our method, respectively, in which words with `\$' ahead mean concept words, and the same applies to subsequent results.}
%     \label{fig:qual}
%     \vspace{-5mm}
% \end{figure*}

\subsection{Comparisons with Existing Methods}

\textbf{Quantitative Evaluation.}
As illustrated in Tab.~\ref{tab:quan-concept} and Tab.~\ref{tab:quan-no_concept}, we assess text-driven video editing results in three aspects. 
Compared with existing methods that extend and fine-tune the Stable Diffusion model, including Tune-A-Video~\cite{wu2023tune}, FateZero~\cite{qi2023fatezero}, RAVE~\cite{kara2024rave}, and MotionDirector~\cite{zhao2023motiondirector}, our approach demonstrates superior inter-frame coherence in terms of the Frame Consistency Metric. 
% compared to other methods that extend and fine-tune the Stable-Diffusion model
% Specifically, we utilize the Frame Consistency metric to evaluate the frame consistency of edited videos. 
To evaluate the consistency of unrelated areas before and after video editing, we employ M-PSNR as a reference metric, and our method achieves the highest score by a large margin. 
Concretely, our method outperforms MotionDirector~\cite{zhao2023motiondirector} by a noticeable 6.98 M-PSNR in editing with concept video. 
This is attributed to our proposed prior supervision mechanism, which effectively reduces the editing noise in non-target areas for both source and concept videos. 
% As evidenced in Fig.~\ref{fig:qual} indicate that the attention supervision mechanism we proposed effectively reduces noise from both the source video and the concept video, thereby enhancing the consistency of non-target regions in the edited results. 
Furthermore, to evaluate the target fidelity in concept and edited videos, 
% in the edited results and the target in the concept video, 
we utilize Concept Consistency as a reference metric, and our method demonstrates greater fidelity compared to others. 

% As illustrated in Table~\ref{tab:quan}, to compare the frame consistency of the video editing results, we use the Frame Consistency metric as a numerical reference. MotionDirector~\cite{zhao2023motiondirector}, based on a pre-trained text-driven video generation model, has an advantage in inter-frame coherence. Our shows an advantage in inter-frame coherence compared to other methods that extend and fine-tune the Stable-Diffusion model, such as Tune-A-Video~\cite{wu2023tune} and FateZero ~\cite{qi2023fatezero}, as well as the RAVE~\cite{kara2024rave} that uses ControlNet \cite{zhang2023adding} to guide Stable-Diffusion for video editing. To compare the consistency of unrelated regions before and after video editing, we use M-PSNR as a reference metric, our method achieves higher scores compared to other methods. As indicated by the results in Figure~\ref{fig:qual}, the attention supervision we proposed can effectively reduce noise from the source video and concept video, and improve the consistency of non-target regions in the edited results. To compare the fidelity between the replaced target in the edited results and the target in the concept video, we use concept consistency as a reference metric, our method achieves higher faithfulness compared to other methods.

\begin{figure}[!ht]
    \centering
    % \vspace{-3mm}
    \includegraphics[width=\linewidth]{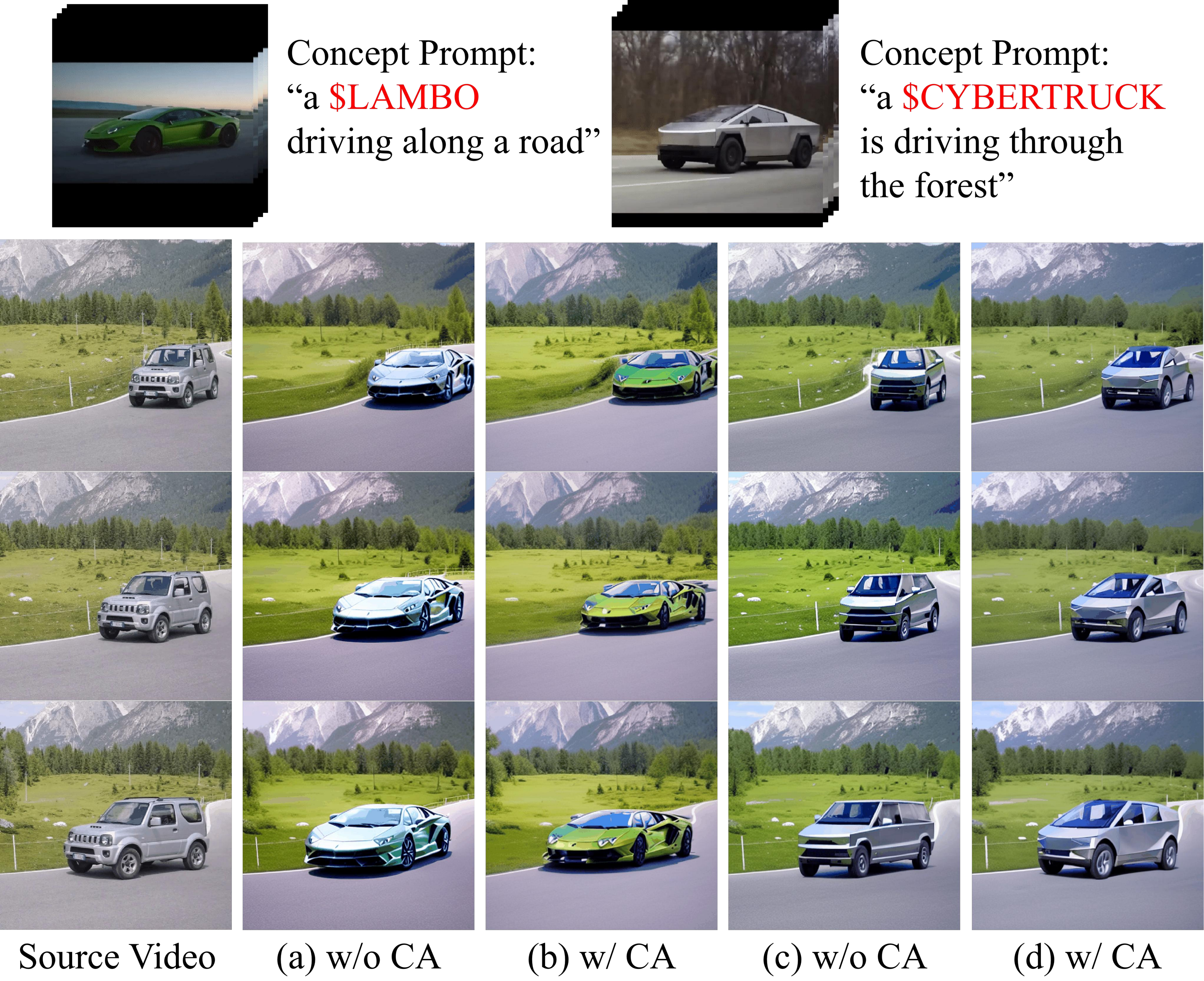}
    % \vspace{-4mm}
    \caption{\textbf{Comparison of whether to use Concept Augmentation (CA) for textual inversion.} Compared the text inversion results without and with concept augmentation for pairs (a), (b): `jeep' $\to$ \textcolor{red}{`\$LAMBO'}; and (c), (d): `jeep' $\to$ \textcolor{red}{`\$CYBERTRUCK'}, respectively, from the same source prompt ``a jeep driving down a curvy road in the countryside".}
    \label{fig:abla-lora}
    % \vspace{-3mm}
\end{figure}

\noindent\textbf{Qualitative Evaluation.}
The visual comparison results of video editing with and without concept video guidance are shown in Fig.~\ref{fig:qual}. 
As can be seen, our method can maintain content consistency in non-target areas before and after video editing with and without concept videos. 
Particularly, when using concept videos, our method can effectively introduce the visual concept from the concept video into the edited video. 
For example, in Fig.~\ref{fig:qual} (Setting I), our method successfully replaces `man' with the concept `\$OPTIMUS', while others either fail to preserve the background or cannot transfer the integral target shape. 
% The visual comparison of our proposed method with Tune-A-Video~\cite{wu2023tune} and FateZero~\cite{qi2023fatezero} for video editing with and without concept video guidance is shown in Fig.~\ref{fig:qual}. 

On the other hand, other approaches commonly face instability in non-target areas of their edited videos. 
Tune-A-Video~\cite{wu2023tune} encounters the issue of dispersed cross-attention between word embeddings and video latents as it fine-tunes the model using only one video-text pair. 
% This leads to instability in non-target areas of edited videos, such as noticeable changes in background color or even content. 
While FateZero~\cite{qi2023fatezero} and RAVE~\cite{kara2024rave} mitigate this issue by manipulating cross-attention matrices or shuffling noise in a zero-shot manner, these methods directly use concepts to drive video editing, which results in compromised non-target area consistency and degraded concept fidelity. 
% mitigates this issue to some extent by partially replacing cross-attention probability matrices in a zero-shot manner, it 
% directly using the concept to drive video editing may affect the consistency between the video before and after editing, 
% RAVE~\cite{kara2024rave} as well, since the concept words may introduce noise information from unrelated regions during inversion. 
MotionDirector~\cite{zhao2023motiondirector} naturally supports extracting targets from concept videos via its trainable spatial path; however, the coupled spatial and temporal paths struggle to provide stable guidance, leading to noticeable inconsistencies in non-target areas. 
% Despite MotionDirector~\cite{zhao2023motiondirector} natively supports introducing targets from concept videos through its trainable spatial path, when using the spatial path to intervene in the temporal path to drive video editing, the lack of effective and stable guidance causes the editing results to be uncontrollably influenced by the spatial path, resulting in low consistency of non-target areas. 
In contrast, our proposed concept-augmented textual inversion and dual prior supervision can effectively maintain content consistency in non-target areas before and after video editing while accurately capturing specific attributes of user-provided concepts. 
% In contrast, the attention supervision we proposed can effectively address both of these issues, maintaining consistency in non-target areas of the video before and after editing while effectively embodying the specific attributes of the target concept.
% \begin{figure}[!htb]
%     \centering
%     \includegraphics[width=1.0\textwidth]{figures/qualitative.pdf}
%     \caption{\textbf{Qualitative comparison of text-driven video editing with and without concept video.} At the top of the figure are the source video frames to be edited and its corresponding descriptive prompt. The rows below show the editing results of the source video using the editing prompt, for \cite{wu2023tune}, \cite{qi2023fatezero}, \cite{zhao2023motiondirector}, \cite{kara2024rave} and our method, respectively.}
%     \label{fig:qual}
% \end{figure}

\subsection{Ablation Study}

\begin{figure}[t]
    \centering
    % \vspace{-3mm}
    \includegraphics[width=\linewidth]{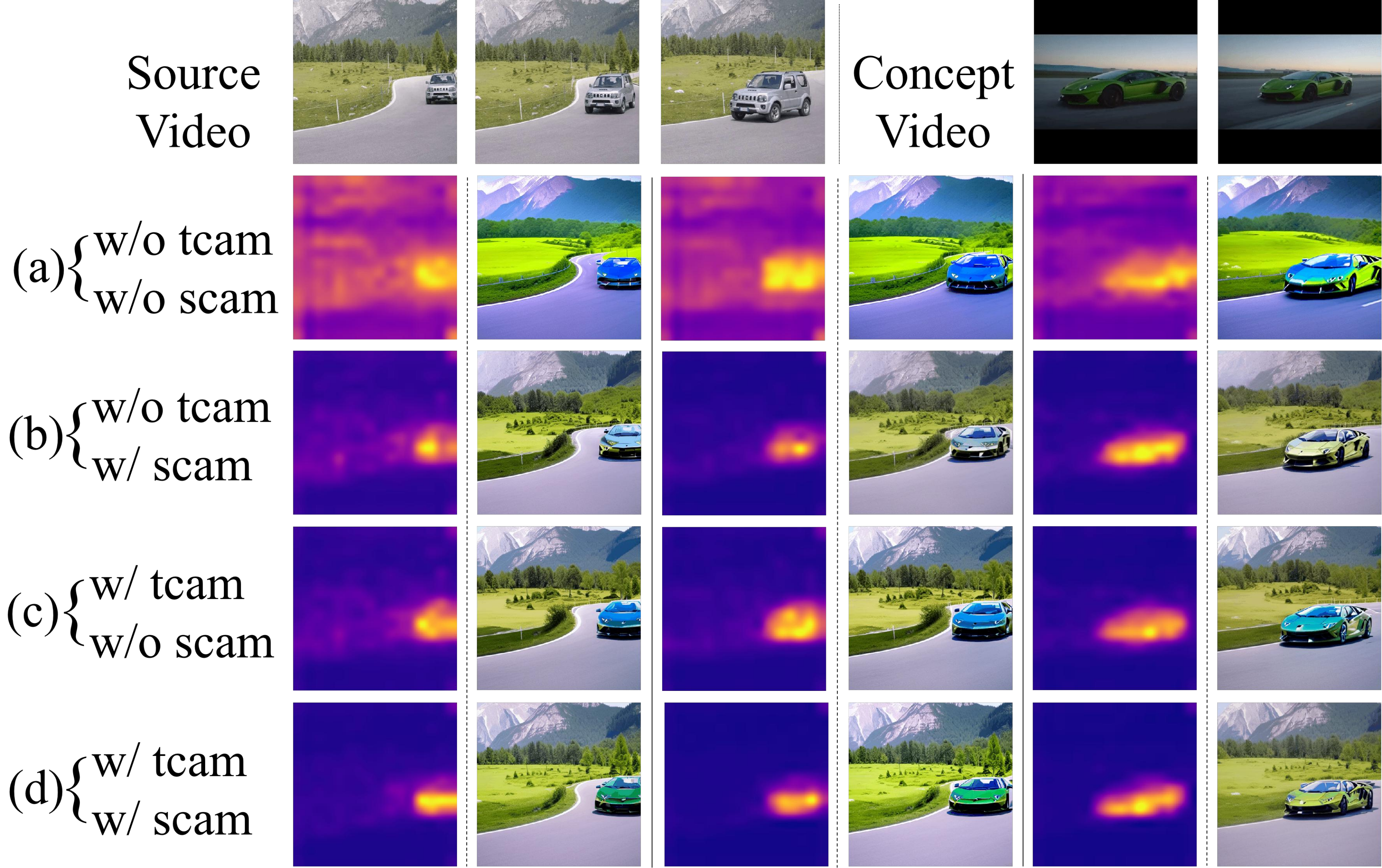}
    % \vspace{-2mm}
    \caption{\textbf{The impact of dual prior supervision.} From the first to the last row, using the editing example in Fig.~\ref{fig:overview}, we compare the average cross-attention maps and the editing results with and without the supervision mechanism of \textit{scam} and \textit{tcam}. Each case contains three pairs, and each pair consists of an average cross-attention map on the left and an edited frame on the right.}
    % \vspace{-3mm}
    \label{fig:abla-full-supvis}
\end{figure}

\textbf{Concept Augmentation Alleviates Under-Fitting of Textual Inversion.}
In this work, we draw on the idea of Textual Inversion (TI) from text-to-image generation and apply it to text-driven video editing to address the embedding of external concept words. However, simply applying TI may lead to under-fitting, resulting in a lack of realism. For instance, in the results shown in Fig.~\ref{fig:abla-lora}(a) and Fig.~\ref{fig:abla-lora}(c), where the keywords `jeep' are altered to `\$LAMBO' and `\$CYBERTRUCK', although some attributes (e.g., shape) of the target concepts are partially retained, the results appear to ``drift" due to insufficient inductive bias. In contrast, the concept-augmented textual inversion (CATI) can effectively capture the color, shape, and other attributes, as demonstrated in Fig.~\ref{fig:abla-lora}(b) and Fig.~\ref{fig:abla-lora}(d). CATI provides more detailed features for editing, significantly improving inversion fidelity.

% As shown in Fig.~\ref{fig:abla-lora}, using Concept Augmented Textual-Inversion can more accurately reflect the color, shape, and other attributes of the target concept in the editing results. Due to the limited latent space of diffusion models and the differences in latent space distribution among various concept videos, Textual Inversion~\cite{gal2022textual} results face the risk of under-fitting, leading to insufficient reflection of the detailed characteristics of the inversion target. However, Textual Inversion augmented with LoRA can effectively enhance the fidelity of details in the inversion results.

\noindent\textbf{Dual Prior Supervision Improves Stability and Fidelity.}
In this work, we propose a Dual Prior Supervision (DPS) strategy, which consists of two main components (See Sec.~\ref{sec:dps}): \textit{scam} loss and \textit{tcam} loss. Both components play crucial roles in maintaining the stability of the target generation. By comparing the attention regions in Fig.~\ref{fig:abla-full-supvis} (a) (w/o \textit{tcam}, w/o \textit{scam}), Fig.~\ref{fig:abla-full-supvis} (b) (w/o \textit{tcam}, w/ \textit{scam}), and Fig.~\ref{fig:abla-full-supvis} (c) (w/ \textit{tcam}, w/o \textit{scam}), we can conclude that both \textit{scam} and \textit{tcam} (Fig.~\ref{fig:abla-full-supvis} (d)) significantly reduce background disturbances and improve stability. However, the generated video results reveal that using either component alone cannot effectively capture attributes of the target object, such as the color of the car. DPS combines both components, not only enhancing the stability of the background in the target results but also capturing the target object's attributes more accurately, thereby improving the fidelity of the edited concept target.

% As shown in Fig.~\ref{fig:abla-full-supvis}, the editing results using both scam and tcam attention supervision mechanisms demonstrate more stable backgrounds and more faithful target concept attributes compared to other supervision mechanisms. Compared to using only scam supervision, our method effectively improves the stability of the background in the editing results by constraining the attention between the concept and irrelevant areas of the source video through tcam. Compared to using only the tcam supervision mechanism, the scam mechanism effectively reduces the possibility of introducing noise from irrelevant areas during tuning with target words, thereby improving the fidelity of the edited concept target.

\begin{figure}[!htb]
    \centering
    \includegraphics[width=\linewidth]{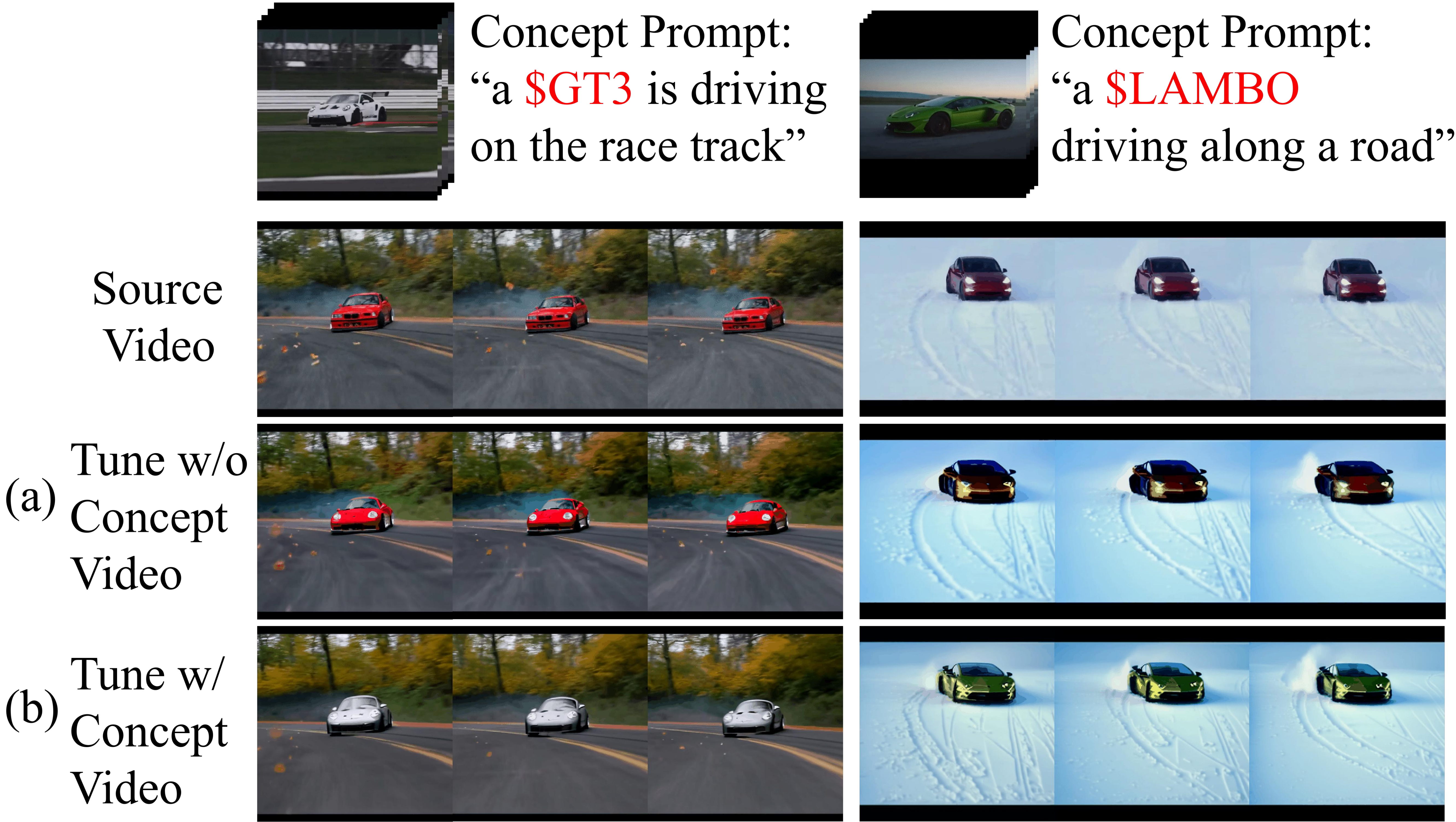}
    % \vspace{-2mm}
    \caption{\textbf{Comparison of whether to tune with the concept video.} Compared the video editing results without and with tuning concept video for the left part: `car' $\to$ \textcolor{red}{`\$GT3'}; and the right part: `car' $\to$ \textcolor{red}{`\$LAMBO'}, from the source prompt ``a car is drifting around a curve road with the background of a forest" and ``a car is drifting in the snow", respectively.}
    \label{fig:abla-tune-concept}
    % \vspace{-5mm}
\end{figure}

\noindent\textbf{Tuning w/ Concept Video Produces Stylized Results.}
Recall that we construct the target videos in this work by templating the concept pairs to make the editing process more flexible. To explore the impact of the concept video in \textbf{Setting I} (Fig.~\ref{fig:qual}), we conduct a simple experiment as shown in Fig.~\ref{fig:abla-tune-concept}. As shown in Fig.~\ref{fig:abla-tune-concept}(a) and Fig.~\ref{fig:abla-tune-concept}(b), tuning models with both concept video and concept prompt produces more stylized videos. The possible explanation lies in that the concept video alleviates the overfitting issue. 
% Note that tuning with only the concept video (w/o the concept prompt) is not viable here, as we cannot analyze the intent without any textual guidance.

% As illustrated in Fig.~\ref{fig:abla-tune-concept}, fine-tuning the video diffusion model with concept videos can effectively improve the faithfulness of consistency between the replacement target and the concept. When fine-tuning the video diffusion model using only a single sample of the source video, the newly added trainable parameters for temporal extension are prone to over-fitting the content in the source video. However, incorporating concept videos into the tuning process effectively mitigates this over-fitting problem.

\section{Limitations and Future Work}
\textbf{Mismatch when Significant Deformation.} 
Although our proposed method effectively mitigates the inconsistency in non-target areas caused by attention dispersion in video editing methods using attention replacement mechanisms, it may struggle when a single concept video guides target replacement in cases of significant deformation in the source video, such as running people. For instance, there may be insufficient detailed correspondences between the internal parts of the replacing and replaced targets during deformation, such as moving arms and legs. Potential solutions include ControlNet~\cite{zhang2023adding}, OpenPose~\cite{8765346} and Sign-D2C~\cite{tang2025discrete} which utilize motion conditions, like human pose or sign language, to guide the video editing process.
% Although our proposed method effectively alleviates the problem of inconsistency in non-target regions caused by attention dispersion in video editing methods based on attention replacement mechanisms, it may fail when using a single concept video to guide target replacement in scenarios where the replaced target undergoes significant deformation in the source video, such as running people. For example, lack of more detailed correspondences between the internal parts in the deformation progress of the replacing and replaced targets, like moving arms and legs. Possible solutions like ControlNet~\cite{zhang2023adding} and OpenPose~\cite{8765346}, utilizing motion conditions such as human pose to guide the video editing process.
\section{Conclusion}
In this paper, we present an improved concept-augmented video editing approach that flexibly produces diverse, stable target videos by leveraging abstract conceptual pairs. Specifically, we introduce Concept-Augmented Textual Inversion (CATI) to capture user-defined target concepts, enabling a plug-and-play, stable diffusion pipeline for more stylized editing. We further propose a Dual Prior Supervision (DPS) mechanism to align cross-attention between source and target prompts, preventing unintended changes in non-target regions. Experimental results show that our method significantly enhances flexibility, consistency, and stability in text-driven video editing.
\section*{Acknowledgments} 
This work is supported by the National Natural Science Foundation of China under Grant No. 62472139, by the Anhui Provincial Natural Science Foundation, China (Grant No. 2408085QF191), the Fundamental Research Funds for the Central Universities (Grants No. JZ2024HGTA0178, JZ2023HGQA0097), by the Open Project Program of the State Key Laboratory of CAD\&CG (Grant No. A2403), Zhejiang University.
\FloatBarrier

\bibliographystyle{named}
\bibliography{ijcai25}

\begin{thebibliography}{}

\bibitem[\protect\citeauthoryear{Avrahami \bgroup \em et al.\egroup }{2022}]{Avrahami_2022_CVPR}
Omri Avrahami, Dani Lischinski, and Ohad Fried.
\newblock Blended diffusion for text-driven editing of natural images.
\newblock In {\em Proceedings of the IEEE/CVF Conference on Computer Vision and Pattern Recognition (CVPR)}, pages 18208--18218, June 2022.

\bibitem[\protect\citeauthoryear{Avrahami \bgroup \em et al.\egroup }{2023}]{avrahami2023blendedlatent}
Omri Avrahami, Ohad Fried, and Dani Lischinski.
\newblock Blended latent diffusion.
\newblock {\em ACM Trans. Graph.}, 42(4), jul 2023.

\bibitem[\protect\citeauthoryear{Bar-Tal \bgroup \em et al.\egroup }{2022}]{bar2022text2live}
Omer Bar-Tal, Dolev Ofri-Amar, Rafail Fridman, Yoni Kasten, and Tali Dekel.
\newblock Text2live: Text-driven layered image and video editing.
\newblock In {\em European conference on computer vision}, pages 707--723. Springer, 2022.

\bibitem[\protect\citeauthoryear{Blattmann \bgroup \em et al.\egroup }{2023}]{blattmann2023stable}
Andreas Blattmann, Tim Dockhorn, Sumith Kulal, Daniel Mendelevitch, Maciej Kilian, Dominik Lorenz, Yam Levi, Zion English, Vikram Voleti, Adam Letts, et~al.
\newblock Stable video diffusion: Scaling latent video diffusion models to large datasets.
\newblock {\em arXiv preprint arXiv:2311.15127}, 2023.

\bibitem[\protect\citeauthoryear{Caelles \bgroup \em et al.\egroup }{2019}]{Caelles_arXiv_2019}
Sergi Caelles, Jordi Pont-Tuset, Federico Perazzi, Alberto Montes, Kevis-Kokitsi Maninis, and Luc {Van Gool}.
\newblock The 2019 davis challenge on vos: Unsupervised multi-object segmentation.
\newblock {\em arXiv:1905.00737}, 2019.

\bibitem[\protect\citeauthoryear{{Cao} \bgroup \em et al.\egroup }{2019}]{8765346}
Z.~{Cao}, G.~{Hidalgo Martinez}, T.~{Simon}, S.~{Wei}, and Y.~A. {Sheikh}.
\newblock Openpose: Realtime multi-person 2d pose estimation using part affinity fields.
\newblock {\em IEEE Transactions on Pattern Analysis and Machine Intelligence}, 2019.

\bibitem[\protect\citeauthoryear{Ceylan \bgroup \em et al.\egroup }{2023}]{ceylan2023pix2video}
Duygu Ceylan, Chun-Hao~P Huang, and Niloy~J Mitra.
\newblock Pix2video: Video editing using image diffusion.
\newblock In {\em Proceedings of the IEEE/CVF International Conference on Computer Vision}, pages 23206--23217, 2023.

\bibitem[\protect\citeauthoryear{Chai \bgroup \em et al.\egroup }{2023}]{chai2023stablevideo}
Wenhao Chai, Xun Guo, Gaoang Wang, and Yan Lu.
\newblock Stablevideo: Text-driven consistency-aware diffusion video editing.
\newblock {\em arXiv preprint arXiv:2308.09592}, 2023.

\bibitem[\protect\citeauthoryear{Cohen \bgroup \em et al.\egroup }{2024}]{cohen2024slicedit}
Nathaniel Cohen, Vladimir Kulikov, Matan Kleiner, Inbar Huberman-Spiegelglas, and Tomer Michaeli.
\newblock Slicedit: Zero-shot video editing with text-to-image diffusion models using spatio-temporal slices.
\newblock {\em arXiv preprint arXiv:2405.12211}, 2024.

\bibitem[\protect\citeauthoryear{Gal \bgroup \em et al.\egroup }{2022}]{gal2022textual}
Rinon Gal, Yuval Alaluf, Yuval Atzmon, Or~Patashnik, Amit~H. Bermano, Gal Chechik, and Daniel Cohen-Or.
\newblock An image is worth one word: Personalizing text-to-image generation using textual inversion, 2022.

\bibitem[\protect\citeauthoryear{Geyer \bgroup \em et al.\egroup }{2023}]{geyer2023tokenflow}
Michal Geyer, Omer Bar-Tal, Shai Bagon, and Tali Dekel.
\newblock Tokenflow: Consistent diffusion features for consistent video editing.
\newblock {\em arXiv preprint arXiv:2307.10373}, 2023.

\bibitem[\protect\citeauthoryear{Hertz \bgroup \em et al.\egroup }{2023}]{hertz2023prompttoprompt}
Amir Hertz, Ron Mokady, Jay Tenenbaum, Kfir Aberman, Yael Pritch, and Daniel Cohen-or.
\newblock Prompt-to-prompt image editing with cross-attention control.
\newblock In {\em The Eleventh International Conference on Learning Representations}, 2023.

\bibitem[\protect\citeauthoryear{Hessel \bgroup \em et al.\egroup }{2021}]{hessel2021clipscore}
Jack Hessel, Ari Holtzman, Maxwell Forbes, Ronan~Le Bras, and Yejin Choi.
\newblock {CLIPScore:} a reference-free evaluation metric for image captioning.
\newblock In {\em EMNLP}, 2021.

\bibitem[\protect\citeauthoryear{Ho \bgroup \em et al.\egroup }{2020}]{ho2020denoising}
Jonathan Ho, Ajay Jain, and Pieter Abbeel.
\newblock Denoising diffusion probabilistic models.
\newblock {\em Advances in neural information processing systems}, 33:6840--6851, 2020.

\bibitem[\protect\citeauthoryear{Ho \bgroup \em et al.\egroup }{2022}]{ho2022imagen}
Jonathan Ho, William Chan, Chitwan Saharia, Jay Whang, Ruiqi Gao, Alexey Gritsenko, Diederik~P Kingma, Ben Poole, Mohammad Norouzi, David~J Fleet, et~al.
\newblock Imagen video: High definition video generation with diffusion models.
\newblock {\em arXiv preprint arXiv:2210.02303}, 2022.

\bibitem[\protect\citeauthoryear{Hong \bgroup \em et al.\egroup }{2022}]{hong2022cogvideo}
Wenyi Hong, Ming Ding, Wendi Zheng, Xinghan Liu, and Jie Tang.
\newblock Cogvideo: Large-scale pretraining for text-to-video generation via transformers.
\newblock {\em arXiv preprint arXiv:2205.15868}, 2022.

\bibitem[\protect\citeauthoryear{Hu \bgroup \em et al.\egroup }{2022}]{hu2022lora}
Edward~J Hu, Yelong Shen, Phillip Wallis, Zeyuan Allen-Zhu, Yuanzhi Li, Shean Wang, Lu~Wang, and Weizhu Chen.
\newblock Lo{RA}: Low-rank adaptation of large language models.
\newblock In {\em International Conference on Learning Representations}, 2022.

\bibitem[\protect\citeauthoryear{Kara \bgroup \em et al.\egroup }{2024}]{kara2024rave}
Ozgur Kara, Bariscan Kurtkaya, Hidir Yesiltepe, James~M. Rehg, and Pinar Yanardag.
\newblock Rave: Randomized noise shuffling for fast and consistent video editing with diffusion models.
\newblock In {\em Proceedings of the IEEE/CVF Conference on Computer Vision and Pattern Recognition}, 2024.

\bibitem[\protect\citeauthoryear{Kasten \bgroup \em et al.\egroup }{2021}]{kasten2021layered}
Yoni Kasten, Dolev Ofri, Oliver Wang, and Tali Dekel.
\newblock Layered neural atlases for consistent video editing.
\newblock {\em ACM Transactions on Graphics (TOG)}, 40(6):1--12, 2021.

\bibitem[\protect\citeauthoryear{Ku \bgroup \em et al.\egroup }{2024}]{ku2024anyv2v}
Max Ku, Cong Wei, Weiming Ren, Harry Yang, and Wenhu Chen.
\newblock Anyv2v: A tuning-free framework for any video-to-video editing tasks.
\newblock {\em arXiv preprint arXiv:2403.14468}, 2024.

\bibitem[\protect\citeauthoryear{Lee \bgroup \em et al.\egroup }{2023}]{lee2023shape}
Yao-Chih Lee, Ji-Ze~Genevieve Jang, Yi-Ting Chen, Elizabeth Qiu, and Jia-Bin Huang.
\newblock Shape-aware text-driven layered video editing.
\newblock In {\em Proceedings of the IEEE/CVF Conference on Computer Vision and Pattern Recognition}, pages 14317--14326, 2023.

\bibitem[\protect\citeauthoryear{Liu \bgroup \em et al.\egroup }{2024}]{liu2024video}
Shaoteng Liu, Yuechen Zhang, Wenbo Li, Zhe Lin, and Jiaya Jia.
\newblock Video-p2p: Video editing with cross-attention control.
\newblock In {\em Proceedings of the IEEE/CVF Conference on Computer Vision and Pattern Recognition}, pages 8599--8608, 2024.

\bibitem[\protect\citeauthoryear{Loshchilov and Hutter}{2017}]{loshchilov2017decoupled}
Ilya Loshchilov and Frank Hutter.
\newblock Decoupled weight decay regularization.
\newblock {\em arXiv preprint arXiv:1711.05101}, 2017.

\bibitem[\protect\citeauthoryear{Minderer \bgroup \em et al.\egroup }{2022}]{minderer2022simple}
Matthias Minderer, Alexey Gritsenko, Austin Stone, Maxim Neumann, Dirk Weissenborn, Alexey Dosovitskiy, Aravindh Mahendran, Anurag Arnab, Mostafa Dehghani, Zhuoran Shen, et~al.
\newblock Simple open-vocabulary object detection.
\newblock In {\em European Conference on Computer Vision}, pages 728--755. Springer, 2022.

\bibitem[\protect\citeauthoryear{Qi \bgroup \em et al.\egroup }{2023}]{qi2023fatezero}
Chenyang Qi, Xiaodong Cun, Yong Zhang, Chenyang Lei, Xintao Wang, Ying Shan, and Qifeng Chen.
\newblock Fatezero: Fusing attentions for zero-shot text-based video editing.
\newblock {\em arXiv:2303.09535}, 2023.

\bibitem[\protect\citeauthoryear{Radford \bgroup \em et al.\egroup }{2021}]{radford2021learning}
Alec Radford, Jong~Wook Kim, Chris Hallacy, Aditya Ramesh, Gabriel Goh, Sandhini Agarwal, Girish Sastry, Amanda Askell, Pamela Mishkin, Jack Clark, et~al.
\newblock Learning transferable visual models from natural language supervision.
\newblock In {\em International conference on machine learning}, pages 8748--8763. PMLR, 2021.

\bibitem[\protect\citeauthoryear{Rombach \bgroup \em et al.\egroup }{2021}]{rombach2021highresolution}
Robin Rombach, Andreas Blattmann, Dominik Lorenz, Patrick Esser, and Björn Ommer.
\newblock High-resolution image synthesis with latent diffusion models, 2021.

\bibitem[\protect\citeauthoryear{Singer \bgroup \em et al.\egroup }{2025}]{singer2025video}
Uriel Singer, Amit Zohar, Yuval Kirstain, Shelly Sheynin, Adam Polyak, Devi Parikh, and Yaniv Taigman.
\newblock Video editing via factorized diffusion distillation.
\newblock In {\em European Conference on Computer Vision}, pages 450--466. Springer, 2025.

\bibitem[\protect\citeauthoryear{Song \bgroup \em et al.\egroup }{2020}]{song2020denoising}
Jiaming Song, Chenlin Meng, and Stefano Ermon.
\newblock Denoising diffusion implicit models.
\newblock {\em arXiv preprint arXiv:2010.02502}, 2020.

\bibitem[\protect\citeauthoryear{Tang \bgroup \em et al.\egroup }{2025}]{tang2025discrete}
Shengeng Tang, Jiayi He, Lechao Cheng, Jingjing Wu, Dan Guo, and Richang Hong.
\newblock Discrete to continuous: Generating smooth transition poses from sign language observation.
\newblock In {\em CVPR}, 2025.

\bibitem[\protect\citeauthoryear{Vaswani}{2017}]{vaswani2017attention}
A~Vaswani.
\newblock Attention is all you need.
\newblock {\em Advances in Neural Information Processing Systems}, 2017.

\bibitem[\protect\citeauthoryear{Wang \bgroup \em et al.\egroup }{2024}]{wang2024videocomposer}
Xiang Wang, Hangjie Yuan, Shiwei Zhang, Dayou Chen, Jiuniu Wang, Yingya Zhang, Yujun Shen, Deli Zhao, and Jingren Zhou.
\newblock Videocomposer: Compositional video synthesis with motion controllability.
\newblock {\em Advances in Neural Information Processing Systems}, 36, 2024.

\bibitem[\protect\citeauthoryear{Wu \bgroup \em et al.\egroup }{2023}]{wu2023tune}
Jay~Zhangjie Wu, Yixiao Ge, Xintao Wang, Stan~Weixian Lei, Yuchao Gu, Yufei Shi, Wynne Hsu, Ying Shan, Xiaohu Qie, and Mike~Zheng Shou.
\newblock Tune-a-video: One-shot tuning of image diffusion models for text-to-video generation.
\newblock In {\em Proceedings of the IEEE/CVF International Conference on Computer Vision}, pages 7623--7633, 2023.

\bibitem[\protect\citeauthoryear{Yang and Tang}{2022}]{yang2022supervised}
Gene-Ping Yang and Hao Tang.
\newblock Supervised attention in sequence-to-sequence models for speech recognition.
\newblock In {\em ICASSP 2022-2022 IEEE International Conference on Acoustics, Speech and Signal Processing (ICASSP)}, pages 7222--7226. IEEE, 2022.

\bibitem[\protect\citeauthoryear{Zhang \bgroup \em et al.\egroup }{2023}]{zhang2023adding}
Lvmin Zhang, Anyi Rao, and Maneesh Agrawala.
\newblock Adding conditional control to text-to-image diffusion models.
\newblock In {\em Proceedings of the IEEE/CVF International Conference on Computer Vision}, pages 3836--3847, 2023.

\bibitem[\protect\citeauthoryear{Zhao \bgroup \em et al.\egroup }{2023}]{zhao2023motiondirector}
Rui Zhao, Yuchao Gu, Jay~Zhangjie Wu, David~Junhao Zhang, Jiawei Liu, Weijia Wu, Jussi Keppo, and Mike~Zheng Shou.
\newblock Motiondirector: Motion customization of text-to-video diffusion models.
\newblock {\em arXiv preprint arXiv:2310.08465}, 2023.

\end{thebibliography}

\end{document}